\documentclass[11pt,a4paper]{article}
\usepackage[hyperref]{acl2021}
\usepackage{times}
\usepackage{latexsym}

\usepackage{amssymb}
\usepackage{microtype}

\aclfinalcopy 
\def\aclpaperid{3609} 

\usepackage{amsmath}
\usepackage{booktabs}
\usepackage{multirow}
\usepackage{soul}
\usepackage{tikz}
\usetikzlibrary{positioning}
\usepackage{hyperref}
\usepackage{footmisc}
\usetikzlibrary{shapes,snakes}
\usepackage[T1]{fontenc}

\newcommand{\eat}[1]{}

\newcommand{\mysecref}[1]{Section~\ref{#1}}
\newcommand{\mytabref}[1]{Table~\ref{#1}}

\newcommand{\myconstraint}[1]{Constraint (\ref{#1})}
\newcommand{\myfigref}[1]{Figure~\ref{#1}}
\newcommand{\myapdxref}[1]{Appendix~\ref{#1}}

\newcommand{\squishlist}{
 \begin{list}{$\bullet$}
  { \setlength{\itemsep}{0pt}
     \setlength{\parsep}{3pt}
     \setlength{\topsep}{3pt}
     \setlength{\partopsep}{0pt}
     \setlength{\leftmargin}{1.5em}
     \setlength{\labelwidth}{1em}
     \setlength{\labelsep}{0.5em} } }
\newcommand{\squishend}{\end{list}}

\newcommand\mycminipage[2] {\begin{minipage}{#1\textwidth}\centering #2\end{minipage}}

\title{{LNN-EL}: A Neuro-Symbolic Approach to Short-text Entity Linking}

\author{
Hang Jiang\textsuperscript{$1$}\Thanks{~Equal contribution; Author Hang Jiang did this work while interning at IBM.}, 
Sairam Gurajada\textsuperscript{$2$}\footnotemark[1], 
Qiuhao Lu\textsuperscript{$3$}, 
Sumit Neelam\textsuperscript{$2$}, 
Lucian Popa\textsuperscript{$2$},
\\ \bf Prithviraj Sen\textsuperscript{$2$}, Yunyao Li\textsuperscript{$2$}, Alexander Gray\textsuperscript{$2$} \\
\textsuperscript{$1$}MIT \hspace{0.5cm} \textsuperscript{$2$}IBM Research \hspace{0.5cm} \textsuperscript{$3$}University of Oregon\\
{\small\texttt{hjian42@mit.edu}, \texttt{\{sairam.gurajada, alexander.gray\}@ibm.com}, \texttt{luqh@cs.uoregon.edu}},\\ {\small \texttt{sumit.neelam@in.ibm.com}, \texttt{\{lpopa,senp,yunyaoli\}@us.ibm.com}}
}

\date{}

\begin{document}
\maketitle

\begin{abstract}
Entity linking (EL), the task of disambiguating mentions in text by linking them to entities in a knowledge graph, is crucial for text understanding, question answering or conversational systems. Entity linking on short text (e.g., single sentence or question) poses particular challenges due to limited context. 
While prior approaches use either heuristics or black-box neural methods, here we propose LNN-EL, a neuro-symbolic approach that combines the advantages of using interpretable rules based on first-order logic with the performance of neural learning. Even though constrained to using rules, LNN-EL performs competitively against SotA black-box neural approaches, with the added benefits of extensibility and transferability. 
In particular, we show that we can easily blend existing rule templates given by a human expert, with multiple types of features (priors, BERT encodings, box embeddings, etc), and even scores resulting from previous EL methods, thus improving on such methods. For instance, on the LC-QuAD-1.0 dataset, we show more than $4$\% increase in F1 score over previous SotA. Finally, we show that the inductive bias offered by using logic results in learned rules that transfer well across datasets, even without fine tuning, while maintaining high accuracy. 
\end{abstract}
\section{Introduction}
\label{sec:intro}
\textit{Entity Linking} (EL) is the task of disambiguating textual \emph{mentions} by linking them to canonical entities provided by a knowledge graph (KG) such as DBpedia, YAGO~\cite{Yago} or Wikidata~\cite{10.1145/2629489}. A large body of existing work deals with EL in the context of longer text (i.e., comprising of multiple sentences)~\cite{bunescu:eacl06}. The general approach is: 1) extract {\em features} measuring some degree of similarity between the textual mention and any one of several candidate entities~\cite{10.1145/1321440.1321475,cucerzan-2007-large,ratinov-etal-2011-local}, followed by 2) the disambiguation step, either heuristics-based (non-learning)~\cite{hoffart-etal-2011-robust,sakor-etal-2019-old,10.1109/MS.2011.122} or learning-based~\cite{10.1145/1321440.1321475,cucerzan-2007-large,ratinov-etal-2011-local,10.1145/2396761.2396832,ganea-hofmann-2017-deep}, to link the mention to an actual entity.

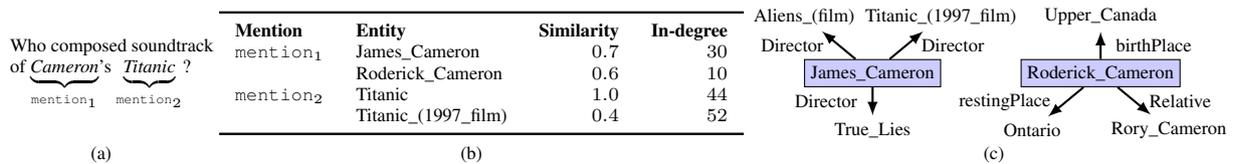
\begin{figure*}
\begin{minipage}{0.17\linewidth}
\scriptsize
\begin{tabbing}
Who composed soundtrack\\
of $\underbrace{\text{{\it Cameron}}}_{\text{{\tt mention}}_1}$'s $\underbrace{\text{{\it Titanic}}}_{\text{{\tt mention}}_2}$?
\end{tabbing}
\end{minipage}
\begin{minipage}{0.43\linewidth}
\scriptsize
\begin{tabular}{llrr}
\toprule
{\bf Mention} & {\bf Entity} & {\bf Similarity} & {\bf In-degree}\\
$\text{{\tt mention}}_1$ & James\_Cameron & $0.7$ & $30$\\
& Roderick\_Cameron & $0.6$ & $10$\\
$\text{{\tt mention}}_2$ & Titanic & $1.0$ & $44$\\
& Titanic\_(1997\_film) & $0.4$ & $52$\\
\bottomrule
\end{tabular}
\end{minipage}
\begin{minipage}{0.4\linewidth}
\scriptsize
\begin{tikzpicture}[>=latex]
\node (jc) at (0,0) [rectangle,draw,fill=blue!20] {James\_Cameron};
\node (tf) at (0.9,0.75) {Titanic\_(1997\_film)};
\node (af) at (-0.9,0.75) {Aliens\_(film)};
\node (tlf) at (0,-0.75) {True\_Lies};
\draw[->,thick] (jc) -- (tf) node[midway,right=3pt]{Director};
\draw[->,thick] (jc) -- (af) node[midway,left=3pt]{Director};
\draw[->,thick] (jc) -- (tlf) node[midway,left=3pt]{Director};

\node (rc) at (3,0) [rectangle,draw,fill=blue!20] {Roderick\_Cameron};
\node (rory) at (3.9,-0.75) {Rory\_Cameron};
\node (o) at (2.1,-0.75) {Ontario};
\node (uc) at (3,0.75) {Upper\_Canada};
\draw[->,thick] (rc) -- (rory) node[midway,right=3pt]{Relative};
\draw[->,thick] (rc) -- (o) node[midway,left=3pt]{restingPlace};
\draw[->,thick] (rc) -- (uc) node[midway,right=3pt]{birthPlace};
\end{tikzpicture}
\end{minipage}
\begin{minipage}{0.15\linewidth}
\scriptsize\centerline{(a)}
\end{minipage}
\begin{minipage}{0.45\linewidth}
\scriptsize\centerline{(b)}
\end{minipage}
\begin{minipage}{0.4\linewidth}
\scriptsize\centerline{(c)}
\end{minipage}
\caption{(a) Question with 2 mentions that need to be disambiguated against DBpedia. (b) For each mention-candidate entity pair, the character-level Jaccard similarity is shown along with the in-degree of the entity in the knowledge graph. (c) (Partial) Ego networks for entities {\tt \small James\_Cameron} and {\tt \small Roderick\_Cameron}.}
\label{fig:introexample}
\vspace*{-4mm}
\end{figure*}

A particular type of entity linking, focused on short text (i.e., a single sentence or question), 
has attracted recent attention due to its relevance for downstream applications such as question answering (e.g., ~\cite{kapanipathi2021leveraging}) and conversational systems. Short-text EL is particularly challenging because the limited context surrounding mentions results in greater ambiguity~\cite{sakor-etal-2019-old}. To address this challenge, one needs to exploit as many features from as many sources of evidence as possible.

Consider the question in \myfigref{fig:introexample}(a), containing {\small \tt mention}$_1$ ({\small \tt Cameron}) and {\small \tt mention}$_2$ ({\small \tt Titanic}).%
\footnote{Note that we assume that mention extraction has already been applied and we are given the textual mentions.} 
DBpedia contains several person entities whose last name matches 
{\small \tt Cameron}. 
Two such entities are shown in \myfigref{fig:overview}(b), {\small \tt James\_Cameron} and {\small \tt Roderick\_Cameron}, 
along with their string similarity scores (in this case, character-level Jaccard similarity) to {\small \tt mention}$_1$. 
In this case, the string similarities are quite close. 
In the absence of reliable discerning information, one can 
employ a prior such as using the more popular candidate entity, as measured by the in-degree of the entity in the KG (see \myfigref{fig:overview}(b)). 
Given the higher in-degree, we can (correctly) link {\small \tt mention}$_1$ to {\small \tt James\_Cameron}.
However, for {\small \tt mention}$_2$, the correct entry is {\small \tt Titanic\_(1997\_film)} as opposed to {\small \tt Titanic} the ship, but it actually has a \emph{lower} string similarity. 
To link to the correct entity, one needs 
to exploit the fact that {\small \tt James\_Cameron} has an edge connecting it to {\small \tt Titanic\_(1997\_film)} in the KG (see ego network on the left in \myfigref{fig:introexample}(c)). 
Linking co-occurring mentions from text to connected entities in the KG is an instance of \textit{collective entity linking}. 
This example provides some intuition as to how priors, local features (string similarity) and collective entity linking can be exploited to overcome the limited context in short-text EL.

While the use of priors, local features and non-local features (for collective linking) has been proposed before \citep{ratinov-etal-2011-local}, our goal in this paper is to provide an {\em extensible} 
framework that can combine any number of such features and more, including contextual embeddings such as BERT encodings~\citep{devlin2019bert} and Query2box embeddings~\cite{DBLP:conf/iclr/RenHL20}, and even the results of 
previously developed neural EL models (e.g., BLINK~\citep{wu2019zero}). 
 
Additionally,  such a framework must 
not only allow for easy inclusion of new sources of evidence but also for \emph{interpretability} of the resulting model \citep{guidotti:acmsurvey18}. An approach that combines disparate features should, at the very least, be able to state, post-training, which features are detrimental 
and which features aid EL performance and under what conditions, in order 
to enable 
actionable insights in the next iteration of model improvement.

\noindent \textbf{Our Approach.}
We propose to use rules in first-order logic (FOL), an interpretable fragment of logic, 
as a glue to combine EL features into a coherent model. Each rule in itself is a disambiguation model capturing specific characteristics of the overall linking. 
While inductive logic programming \citep{muggleton:ilpw96} and statistical relational learning \citep{getoor:book} 
have for long focused on learning FOL rules from labeled data, more recent approaches based on neuro-symbolic AI have led to impressive advances. 
In this work,
we start with an input set of rule templates (given by an expert or available as a library), and learn the parameters of these rules (namely, the thresholds of the various similarity predicates as well as the weights of the predicates that appear in the rules), based on a labeled dataset. We use logical neural networks (LNN)~\cite{riegel2020logical}, a powerful neuro-symbolic AI approach based on real-valued logic that employs neural networks to learn the parameters of the rules. 
Learning of the rule templates themselves will be the focus of future work.

{\bf Summary of contributions} 
\squishlist
\item We propose, to the best of our knowledge, the first neuro-symbolic method for entity linking (coined ``\textit{LNN-EL}") that provides a principled approach to learning EL rules.
\item Our approach is extensible and can combine disparate types of local and global features as well as results of prior black-box neural methods, thus building on top of such approaches.
\item Our approach produces interpretable rules that humans can inspect toward actionable insights.
\item We evaluate our approach on three benchmark datasets and show competitive
(or better) performance with SotA black-box neural approaches (e.g., BLINK~\citep{wu2019zero}) even though 
we are 
constrained on using rules. 
\item By leveraging rules, the learned model shows a desirable {\em transferability} property: it performs well not only on the dataset on which it was trained, but also on other datasets from the same domain without further training.

\squishend

\section{Related Work}
\noindent \textbf{Entity Linking Models.} Entity Linking is a well-studied problem in NLP, especially for long text. Approaches such as 
~\cite{bunescu:eacl06, ratinov-etal-2011-local, sil-etal-2012-linking, hoffart-etal-2011-robust, 6823700} use a myriad of classical ML and deep learning models to combine priors, local and global features. These techniques, in general, can be applied to short text, but the lack of sufficient context
may render them ineffective. 
The recently proposed BLINK~\cite{logeswaran-etal-2019-zero,wu2019zero} uses powerful transformer-based encoder architectures trained on massive amounts of data (such as Wikipedia, Wikia) to achieve SotA performance on entity disambiguation tasks, and is shown to be especially effective in zero-shot settings. 
BLINK is quite effective on short text (as observed in our findings); in our approach, we use BLINK both as a baseline and as a component that is combined in larger rules.

For short-text EL, some prior works~\cite{sakor-etal-2019-old, 10.1109/MS.2011.122,mendes2011dbpedia} address the joint problem of mention detection and linking, with primary focus on identifying mention spans, while linking is done via heuristic methods without learning. \cite{sakor-etal-2019-old} also jointly extracts relation spans which aide in overall linking performance. The recent ELQ~\cite{li2020efficient} extends BLINK to jointly learn mention detection and linking. In contrast, we focus solely on linking and take a 
different strategy  
based on combining logic rules with learning. 
This facilitates a principled way combining multiple types of EL features 
with interpretability and learning using promising gradient-based techniques.

\smallskip
\noindent \textbf{Rule-based Learning.} FOL rules and learning have been successfully applied in some NLP tasks and also other domains.  Of these, the task that is closest to ours is entity resolution (ER), which is the task of linking two entities across two structured datasets. 
In this context, works like ~\cite{Chaudhuri2007ExampledrivenDO,10.1145/1807167.1807252,10.14778/2350229.2350263, 10.1145/2452376.2452440} use FOL rules for ER. Approaches such as ~\cite{10.1109/ICDM.2006.65, Pujara2016GenericSR} induce probabilistic rules using MLNs~\cite{10.1007/s10994-006-5833-1} and  PSL~\cite{10.5555/3122009.3176853}, respectively. 
None of these approaches 
use any recent advances in neural-based learning; moreover, they are 
focused on 
entity resolution, which is a related task but distinct from short-text EL.
\section{Preliminaries}
\subsection{Entity Linking.}
\label{sec:el-definition}
Given text $T$, a set $M=\{m_1, m_2, ...\}$ of mentions, where each $m_i$ is contained in $T$, and a knowledge graph (KG) comprising of a set $\mathcal{E}$ of entities, entity linking is a many-to-one function that links each mention $m_i\in M$ to an entity $e_{ij} \in C_i$, where $C_i\subseteq \mathcal{E}$ is a subset of relevant candidates for mention $m_i$. More generally, we formulate the problem as a ranking of the candidates in $C_i$ so that the ``correct" entity for $m_i$ is ranked highest. 
Following existing approaches(e.g. ~\cite{sakor-etal-2019-old,wu2019zero}, we use off-the-shelf lookup tools such as DBpedia lookup\footnote{\url{https://lookup.dbpedia.org/}} to retrieve top-100 candidates for each mention. While this service is specific to DBpedia, we assume that similar services 
exist or can be implemented on top of other KGs. 

\subsection{Logical Neural Networks}
\label{sec:lnn-prelims}

Fueled by the rise in complexity of deep learning, recently there has been a push towards learning interpretable models \citep{guidotti:acmsurvey18,danilevsky-etal-2020-survey}. 
While linear classifiers, decision lists/trees may also be considered interpretable, rules expressed in first-order logic (FOL) form a much more powerful, closed language that offer semantics clear enough for human interpretation and a larger range of operators facilitating the expression of richer models. To learn these rules, neuro-symbolic AI typically substitutes conjunctions (disjunctions) with differentiable $t$-norms ($t$-conorms) \citep{esteva:fuzzysetssyst01}. However, since these norms do not have any learnable parameters (more details in \myapdxref{apdx:lnn}), 
their behavior cannot be adjusted, thus limiting their ability to model well the data.

\begin{figure}
    \vspace*{-4mm}
    \includegraphics[width=\linewidth]{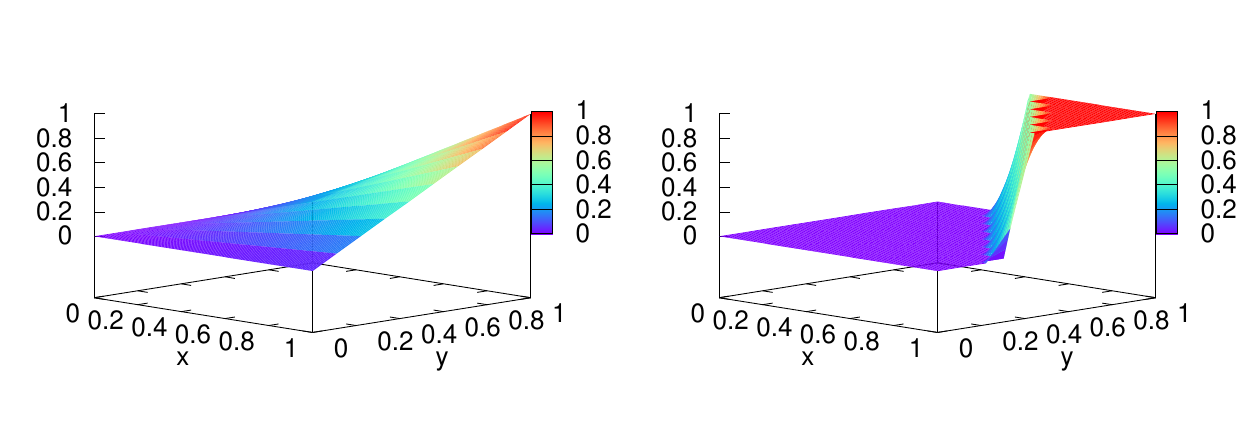}
    \begin{minipage}{0.49\linewidth}
    \scriptsize
    \end{minipage}
    \begin{minipage}{0.49\linewidth}
    \scriptsize
    \end{minipage}
    \vspace*{-6mm}
    \caption{(left) Product $t$-norm. (right) LNN-$\wedge$ ($\alpha=0.7$).}
    \label{fig:lnnand}
    \vspace*{-5mm}
\end{figure}

In contrast, logical neural networks (LNN) \citep{riegel2020logical} offer operators that include parameters, thus allowing to better learn from the data. To maintain the crisp semantics of FOL, LNNs enforce constraints when learning operators such as conjunction. Concretely, LNN-$\wedge$ is expressed as:

\vspace*{-4mm}
{\small 
\begin{align}
\nonumber \max(0, \min(&1, \beta - w_1(1-x) - w_2(1-y)))\\
\label{eqn:hihi}\text{subject to: }&\beta - (1-\alpha)(w_1 + w_2) \geq \alpha\\
\label{eqn:lohi}&\beta - \alpha w_1 \leq 1-\alpha\\
\label{eqn:hilo}&\beta - \alpha w_2 \leq 1-\alpha\\
\nonumber &w_1,w_2 \geq 0
\end{align}
}
where $\beta, w_1, w_2$ are \emph{learnable} parameters, $x,y \in [0,1]$ are inputs and $\alpha \in [\frac{1}{2},1]$ is a hyperparameter. Note that $\max(0, \min(1, \cdot))$ clamps the output of LNN-$\wedge$ between $0$ and $1$ regardless of $\beta, w_1, w_2, x,$ and $y$. The more interesting aspects are in the constraints. 
While Boolean conjunction only returns $1$ or {\tt true} when both inputs are $1$, LNNs relax this condition by 
using $\alpha$ as a proxy for 1 (and conversely, $1 - \alpha$ as a proxy for 0). In particular, 
\myconstraint{eqn:hihi}
forces the output of LNN-$\wedge$ to be greater than $\alpha$ when both inputs are greater than $\alpha$. 
Similarly, 
Constraints (2) and (3) 
constrain the behavior of LNN-$\wedge$ when one input is low and the other is high. For instance, 
\myconstraint{eqn:lohi}
forces the output of LNN-$\wedge$ to be less than $1-\alpha$ for $y=1$ and $x \leq 1-\alpha$. This formulation allows for unconstrained learning when 
$x,y \in [1-\alpha, \alpha]$. 
By 
changing $\alpha$ a user can control how much learning to enable (increase to make region of unconstrained learning wider or decrease 
for 
the opposite). \myfigref{fig:lnnand} depicts product $t$-norm and LNN-$\wedge$ ($\alpha=0.7$). While the former increases slowly with increasing $x,y$, LNN-$\wedge$ produces a high output when both inputs are $\geq \alpha$ and stays high thereafter, thus 
closely modeling Boolean conjunction semantics.

In case the application requires even more degrees of freedom, the hard constraints 
(1), (2) and (3) 
can be relaxed via the inclusion of slacks:

{\small 
\smallskip
\noindent\begin{tikzpicture}

\node at (0,0) (a) {
\begin{minipage}{0.47\textwidth}
\begin{align}
\nonumber \max(0, \min&(1, \beta - w_1(1-x) - w_2(1-y)))\\
\nonumber\text{subject to: }&\beta - (1-\alpha)(w_1 + w_2) + \Delta \geq \alpha\\
\nonumber&\beta - \alpha w_1 \leq 1-\alpha + \delta_1\\
\nonumber&\beta - \alpha w_2 \leq 1-\alpha + \delta_2\\
\nonumber &w_1,w_2,\delta_1,\delta_2,\Delta \geq 0
\end{align}
\end{minipage}};
\node[above=-6.5mm of a] {LNN-$\wedge(x,y)=$};
\end{tikzpicture}
}
\noindent 
where $\delta_1, \delta_2,$ and $\Delta$ denote slack variables. If any of 
Constraints (1), (2) and (3) 
in LNN-$\wedge$ are unsatisfied then slacks help correct the direction of the inequality without putting pressure on parameters $w_1, w_2,$ and $\beta$ during training. For the rest of the paper, by LNN-$\wedge$ we refer to the above formulation. LNN negation is a pass-through operator: $\text{LNN-}\neg(x) = 1 - x$, and LNN disjunction is defined in terms of LNN-$\wedge$:

\vspace*{-4mm}
{\small 
\begin{align*}
\text{LNN-}\!\vee\!(x, y) = 1-\text{LNN-}\!\wedge\!(1-x,1-y)
\end{align*}
}
While vanilla backpropagation cannot handle linear inequality constraints such as 
\myconstraint{eqn:hihi}, 
specialized learning algorithms are available within the LNN framework. 
For more details,
please check \citet{riegel2020logical}. 
\section{LNN-EL}
\begin{figure*}[t]
\vspace*{-1mm}
    \centering
    \resizebox{0.8\textwidth}{0.3\textwidth}{
  \begin{tikzpicture}

    \node[draw, inner sep=2mm] at (-5,-1.2) (t) {Text $T$};
    \node[draw, inner sep=2mm, below=4mm of t] (kgr) {KG Resources};
    \node[draw, inner sep=2mm, below=4mm of kgr] (ff) {Feature functions $\mathcal{F}$};
    
\draw[fill=black!0, rounded corners] (-2.2,1.6) rectangle (2.2,-1);
\node at (0,0) (ld) {\begin{minipage}{0.4\textwidth}    
      \begin{equation*}
        m_i, \begin{bmatrix}
             e_{i1}, l_{i1}\\
             e_{i2}, l_{i2}\\
             \vdots
             \end{bmatrix}  
      \end{equation*} 
  \end{minipage}   
};
\node[above=0mm of ld] {{\bf Labeled data} $m_i,[C_i, L_i]$};

\draw[->,line width=0.8mm] (0,-1.1) -- (0,-1.7);

\draw[fill=blue!20, rounded corners] (-1.8,-1.8) rectangle (1.8,-2.9);
\node[below=8.3mm of ld] (fg) {\begin{minipage}{0.2\textwidth}
    \centering
    Feature\\ Generation
    \end{minipage}
 };

\draw[->,line width=0.8mm] (0,-3.0) -- (0,-3.6);

\draw[fill=black!0, rounded corners] (-2.6,-3.7) rectangle (2.6,-6.6);
  \node[below=19mm of fg] (lgd) {\begin{minipage}{0.4\textwidth}
      \begin{equation*}
        m_i, \begin{bmatrix}
             e_{i1},[f_1,f_2,\ldots]_{i1}, l_{i1}\\
             e_{i2},[f_1,f_2,\ldots]_{i2}, l_{i2}\\
             \vdots
             \end{bmatrix}  
      \end{equation*} 
  \end{minipage}   
};

\node[above=-1mm of lgd] {\mycminipage{0.4}{{\bf  Labeled data with features}\\ $m_i,[C_i, F_i, L_i]$}};

\node at (7.4,-4) (lnnmodel) {\mycminipage{0.5}{
    \tikz{
      \node[fill=blue!20,draw, rounded corners] at (0,0) (and1) {LNN-$\wedge$};
      \node[draw, circle,above=6mm of and1,inner sep=0.1mm] (ft2) {\mycminipage{0.05}{\tikz{
          \draw (0,0) -- (0.12,0);
          \draw (0.12,0) -- (0.17,0.17);

        }}};
       \node[above left=-0.4mm of ft2] {$\theta_2$};
       \node[draw, rounded corners, above=6mm of ft2] (f2) {$f_2$};
      \node[draw, circle,left=6mm of ft2,inner sep=0.1mm] (ft1) {\mycminipage{0.05}{\tikz{
          \draw (0,0) -- (0.12,0);
          \draw (0.12,0) -- (0.17,0.17);
          
        }}};
    \node[above left=-0.4mm of ft1] {$\theta_1$};
    \node[draw, rounded corners, above=6mm of ft1] (f1) {$f_1$};
      \node[draw, circle,right=6mm of ft2,inner sep=0.1mm] (ft3) {\mycminipage{0.05}{\tikz{
          \draw (0,0) -- (0.12,0);
          \draw (0.12,0) -- (0.17,0.17);
          
        }}};
    \node[above left=-0.4mm of ft3] {$\theta_3$};
    \node[draw, rounded corners, above=6mm of ft3] (f3) {$f_3$};
    \draw[->,line width=0.3mm] (f1.south) -- (ft1.north);
    \draw[->,line width=0.3mm] (f2.south) -- (ft2.north);
    \draw[->,line width=0.3mm] (f3.south) -- (ft3.north);
       \draw[->,line width=0.3mm] (ft1.300) -- node[left]{$fw_1$} (and1.150);
       \draw[->,line width=0.3mm] (ft2.south) -- node[left=-2mm]{ $fw_2$} (and1.90);
       \draw[->,line width=0.3mm] (ft3.240) -- node[right]{$fw_3$} (and1.30);
    \node[right=8mm of and1] (d2) {};
    \node[fill=blue!20,draw, rounded corners, right=8mm of d2] (and2) {LNN-$\wedge$};
    \node[above=6.8mm of and2] (d1) {};
    \node[draw, circle,left=4mm of d1,inner sep=0.1mm] (ft4) {\mycminipage{0.05}{\tikz{
          \draw (0,0) -- (0.12,0);
          \draw (0.12,0) -- (0.17,0.17);
          
        }}};
    \node[above left=-0.4mm of ft4] {$\theta_4$};
    \node[draw, rounded corners, above=6mm of ft4] (f4) {$f_1$};
    \node[draw, circle,right=4mm of d1,inner sep=0.1mm] (ft5) {\mycminipage{0.05}{\tikz{
          \draw (0,0) -- (0.12,0);
          \draw (0.12,0) -- (0.17,0.17);
          
        }}};
    \node[above left=-0.4mm of ft5] {$\theta_5$};
    \node[draw, rounded corners, above=6mm of ft5] (f5) {$f_4$};
    
       \draw[->,line width=0.3mm] (ft4.300) -- node[left]{$fw_4$} (and2.120);
       \draw[->,line width=0.3mm] (ft5.240) -- node[right]{$fw_5$} (and2.50);

       \draw[->,line width=0.3mm] (f4.south) -- (ft4.north);
       \draw[->,line width=0.3mm] (f5.south) -- (ft5.north);
       
       \node[fill=blue!20, draw, rounded corners,below=8mm of d2] (or) {LNN-$\vee$};
       \draw[->,line width=0.3mm] (and1.270) -- node[left=1mm]{$rw_1$} (or.150);
       \draw[->,line width=0.3mm] (and2.270) -- node[right=1mm]{$rw_2$} (or.30);

       \node[above=35mm of or] {\bf LNN Reformulation of EL Algorithm};

     }}};

 \draw (4,-1.6) rectangle (10.8,-6.6);
 \draw[->,line width=0.8mm] (2.9,-5.0) -- (3.7,-5.0);

 \draw[fill=black!0, rounded corners] (3.1,2.1) rectangle (11.5,-1);
 \node at (7.3,0.6) (ela) {\mycminipage{0.8}{
     \begin{align}
       R_1(m_i,e_{ij})\leftarrow&~~f_1(m_i,e_{ij}) > \theta_1 \wedge f_2(m_i,e_{ij}) > \theta_2\notag\\
       & \wedge f_3(m_i,e_{ij}) > \theta_3 \notag\\
       &~~~~~~~~~~~~~~\vee \notag\\  
       R_2(m_i,e_{ij})\leftarrow&~~f_1(m_i,e_{ij}) > \theta_4 \wedge f_4(m_i,e_{ij}) > \theta_5\notag
     \end{align}     
  }};

\node[above=-6mm of ela] {\bf User provided EL Algorithm};
\draw[->,line width=0.8mm] (7.3,-1.1) -- (7.3,-1.4);
\draw[->,line width=0.8mm] (11.2,-5.0) -- (12,-5.0);
\node at (14,-4.8) (op) {\mycminipage{0.4}{
    \begin{equation*}
        m_i, \begin{bmatrix}
             e_{i1}, s(m_i,e_{ij})\\
             e_{i2}, s(m_i,e_{i2})\\
             \vdots
             \end{bmatrix}  
      \end{equation*} 
    }};

\node[above=-3mm of op] {\bf Final scores};

  \draw[->, line width=0.3mm] (t.east) -- (-2.5,-1.2) -- (-2.5,-2) -- (-1.8,-2);
  \draw[->,line width=0.3mm] (kgr.east) -- (-1.8,-2.27);
  \draw[->,line width=0.3mm] (ff.east) -- (-2.5,-3.35) -- (-2.5,-2.7) -- (-1.8,-2.7);
  
\node[dotted,draw] at (14.2,-2) (note) {
\mycminipage{0.3}{
{\bf Learnable parameters:}
\vspace{-4mm}
\begin{align*}
\theta_i & \text{-- feature thresholds},\notag\\
fw_i& \text{-- feature weights},\notag\\
rw_i& \text{-- rule weights}\notag
\end{align*}
}
};
\end{tikzpicture}
}
    \vspace*{-2mm}
    \caption{Overview of our approach}
    \label{fig:overview}
    \vspace*{-3mm}
\end{figure*}
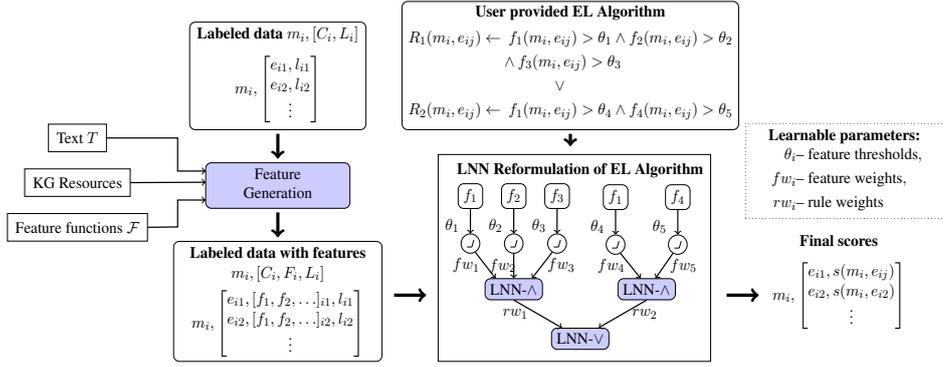
 An overview of our neuro-symbolic approach for entity linking is depicted in Figure~\ref{fig:overview}. We next discuss the details about feature generation component that generates features using a catalogue of feature functions (\mysecref{sec:feature-functions}) followed by proposed model that does neuro-symbolic learning over user provided EL algorithm in \mysecref{sec:lnn-framework}.

Given the input text $T$, together with 
labeled data in the form $(m_i, C_i, L_i)$, where $m_i\in M$ is a mention in $T$, $C_i$ is a 
list of candidate entities $e_{ij}$ (drawn from lookup services\footnote{\url{https://lookup.dbpedia.org}}) for the mention $m_i$, and where 
each $l_{ij}\in L_i$ denotes a link/not-link label for the 
pair $(m_i, e_{ij})$. 
The first step is to generate a set  
$F_{ij} = \{ f_k(m_i,e_{ij})\}$ of features for each pair $(m_i, e_{ij})$, 
where $f_{k}$ is a feature function drawn from a catalog $\mathcal{F}$ of user provided functions.

\begin{table}[t]
\footnotesize
\setlength{\tabcolsep}{5pt}
    \centering
    \begin{tabular}{ll}
         \toprule
         {\bf Features}& {\bf Description}  \\
         \midrule
         Name & $sim(m_i, e_{ij})$, where $sim$ is a general\\ 
         & purpose string similarity function such as\\
         & Jaccard ($jacc$), JaroWinkler ($jw$), \\
         & Levenshtein ($lev$), Partial Ratio ($pr$), etc. \\
         \midrule
         Context &$Ctx(m_i,e_{ij})$\\
         &$=\sum_{\substack{m_k \in M\setminus \{m_i\}}}pr(m_k,e_{ij}.\textsf{desc})$\\
         & where $m_k$ is a mention in the context of $m_i$\\
         \midrule
         Type&$Type(m_i, e_{ij})$\\
         &$=\begin{cases}
        1& \text{if } m_i.\textsf{type} \in e_{ij}.\textsf{dom}\\
        0,              & \text{otherwise}
        \end{cases}$
         \\
         &where $m_i.\textsf{type}$ is the type of the mention \\
         &and $e_{ij}.\textsf{dom}$ is the set of domains  \\
         \midrule
         Entity & $Prom(e_{ij})=indegree(e_{ij})$,\\
         Prominence& i.e., number of links pointing to entity $e_{ij}$ \\
         \bottomrule
    \end{tabular}
    \caption{Non-embedding based feature functions.}
    \label{tab:non-embedding-features}
    \vspace*{-5mm}
\end{table}

\subsection{Feature Functions}
\label{sec:feature-functions}
Our collection of feature functions include both non-embedding and embedding based 
functions. 

\smallskip
\noindent \textbf{Non-embedding based.} We include here a multitude of functions (see \mytabref{tab:non-embedding-features}) that measure the similarity between the mention $m_i$ and the candidate entity $e_{ij}$ based on multiple types of scores.

\smallskip
\textbf{{\em Name:}} 
a set of general purpose string similarity functions\footnote{\url{pypi.org/project/py-stringmatching}} such as {\em Jaccard}, {\em Jaro Winkler}, {\em Levenshtein}, {\em Partial Ratio}, etc. are used to compute the similarity between $m_i$ and $e_{ij}$'s name.

\smallskip
\textbf{{\em Context:}} 
aggregated similarity of $m_i$'s context to the description of $e_{ij}$. Here, we consider the list of all other mentions $m_k\in M$ ($k\neq i$) as $m_i$'s context, together with $e_{ij}$'s textual description obtained using KG resources\footnote{\url{{dbpedia.org/sparql}}\label{fn:sparql-endpoint}}. The 
exact formula we use 
is shown in \mytabref{tab:non-embedding-features}, 
where {\em Partial Ratio}($pr$) measures the similarity between each context mention and the description. 
({\em Partial Ratio} computes the maximum similarity between a short input string and substrings of a second, longer string.)  
For normalizing the final score, we apply a min-max rescaling over all entities $e_{ij}\in C_i$.

\smallskip
\textbf{{\em Type:}} 
the overlap similarity of mention $m_i$'s type to $e_{ij}$'s domain (class) set, similar to the domain-entity coherence score proposed in~\cite{Nguyen2014AIDAlightHN}. Unlike in ~\cite{Nguyen2014AIDAlightHN}, instead of using a single type for all mentions in $M$, we obtain type information for each mention $m_i$ using a trained BERT-based entity type detection model. We use KG resources~\footref{fn:sparql-endpoint} to obtain $e_{ij}$'s domain set, similar to Context similarity.

\smallskip
\textbf{
{\em Entity Prominence:}} measure the prominence of entity $e_{ij}$ as the number of entities that link to $e_{ij}$ in target KG, i.e., $indegree(e_{ij})$. Similar to Context score normalization, we apply min-max rescaling over all entities $e_{ij}\in C_i$.

\medskip
\noindent \textbf{Embedding based.} We also employ a suite of pre-trained or custom trained neural language models to compute the similarity of $m_i$ and 
$e_{ij}$. 

\smallskip
\textbf{{\em Pre-trained Embedding Models.} }These include SpaCy's semantic similarity\footnote{\url{spacy.io/usage/vectors-similarity}} function that uses Glove~\cite{pennington2014glove} trained on Common Crawl.  In addition to SpaCy, we 
also use 
scores from an entity linking system such as BLINK~\cite{wu2019zero} (a state-of-the-art entity linking model) as a feature function in our system.

\smallskip
\textbf{{\em BERT Embeddings.}} To further explore the semantics of the context in $T$ and the inherent structure of the target KG, we incorporate an embedding-based similarity by training a mini entity linking model without any aforementioned prior information. We first tag the input text $T$ with a special token $\rm [MENT]$ to indicate the position of mention $m_i$, and then encode $T$ with BERT, i.e., $\mathbf{m_i}=\text{BERT}(m_i,T)$. Each candidate $e_{ij}$ is encoded with a pre-trained graph embedding Wiki2Vec~\cite{yamada2020wikipedia2vec}, i.e., $\mathbf{e_{ij}}=\text{Wiki2Vec}(e_{ij})$. The candidates are ranked in order of the cosine similarity to $\mathbf{m_i}$, i.e., $Sim_{cos}(\mathbf{m_i},\mathbf{e_{ij}})$. The mini EL model is optimized with margin ranking loss so that the correct candidate is ranked higher.

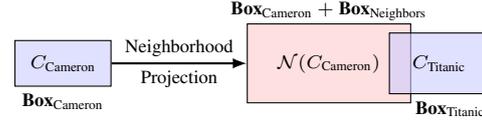
\begin{figure}
    \centering
    \scriptsize
    \begin{tikzpicture}[>=latex]
    \node (cl) at (0,-0.5) {$\textbf{Box}_{\text{Cameron}}$};
    \node (nl) at (3.5,0.7) {$\textbf{Box}_{\text{Cameron}}+\textbf{Box}_{\text{Neighbors}}$};
    \node (tl) at (5.1,-0.6) {$\textbf{Box}_{\text{Titanic}}$};
    \node (c) at (0,0) [rectangle,draw,fill=blue!20,inner sep=2mm,fill opacity=.6,text opacity=1] {$C_{\text{Cameron}}$};
    \node (n) at (3.5,0) [rectangle,draw,fill=red!20,inner sep=4mm,fill opacity=.6,text opacity=1] {$\mathcal{N}(C_\text{Cameron})$};
    \draw[->,thick] (c) to node[above]{Neighborhood} node[below]{Projection} (n);
    \node (t) at (4.95,0) [rectangle,draw,fill=blue!20,inner sep=3mm,fill opacity=.6,text opacity=1] {$C_{\text{Titanic}}$};
    \end{tikzpicture}
    \vspace*{-4mm}
    \caption{Candidates for linking the `Titanic' mention appear in the intersection of the two boxes.}
    \label{boxmap}
    \vspace*{-5mm}
\end{figure}

\smallskip
\textbf{{\em BERT with Box Embeddings.}} 
While features such as {\it Context} (see \mytabref{tab:non-embedding-features}) can exploit other mentions appearing within the same piece of text, 
they only do so via textual similarity. 
A more powerful method is to jointly disambiguate 
the mentions in text to the actual entities in the KG, thus exploiting the structural context 
in 
the 
KG.
Intuitively, the simultaneous linking of co-occurring mentions in text to related entities in the KG is a way to reinforce the links for each individual mention. 
To this end, 
we adapt the recent 
Query2Box~\cite{DBLP:conf/iclr/RenHL20}, whose goal is to answer FOL queries over a KG. 
The main idea there is to represent sets of entities 
(i.e., queries) 
as contiguous regions in embedded space (e.g., axis-parallel hyper-rectangles or boxes), 
thus reducing logical operations to geometric operations (e.g., intersection). 

Since Query2Box assumes a well-formed query as input, one complication in directly applying it to our setting is that we lack the information necessary to form such an FOL query. For instance, in the example from \mysecref{sec:intro}, while we may assume 
that the correct entities for our {\small \tt Cameron} and {\small \tt Titanic} mentions are connected in the KG, we do not know \emph{how} these are connected, i.e., via which relation.
To circumvent this challenge, we 
introduce 
a special \emph{neighborhood} relation $\mathcal{N}$, 
such that 
$v \in \mathcal{N}(u)$ 
whenever 
there is 
some KG relation from entity $u$ to entity $v$.
We next define two box operations:

\vspace*{-4mm}
{\small
\begin{align*}
    & \text{\textbf{Box}}(C_i) = \{\textbf{v} | \min( \{\mathbf{e_{ij}} | e_{ij} \in C_i\} ) \\
    &   \hspace*{1.5cm}~\quad\quad\preceq \textbf{v} \preceq \max( \{\mathbf{e_{ij}} | e_{ij} \in C_i\} ) \}\\
    & \text{\textbf{Box}}(\mathcal{N}(C_i)) = \text{\textbf{Box}}(C_i) + \text{\textbf{Box}}_{\mathcal{N}}
\end{align*}
}
The first operation represents mention $m_i$ as a box, by taking the smallest box that contains 
the set $C_i$ of candidate entities 
for $m_i$.
This can be achieved by computing the dimension-wise minimum (maximum) of all entity embeddings in $C_i$ to obtain the lower-left (upper-right) corner of the resulting box. 
The second operation 
takes $m_i$'s box 
and produces 
the box containing its neighbors in the KG. 
Query2Box achieves this by representing $\text{\textbf{Box}}_{\mathcal{N}}$ via a center vector $\psi$ 
and offset vector $\omega$, both of which are learned parameters. 
The box of neighbors is then obtained by translating the center of $m_i$'s box by $\psi$ and adding the offset $\omega$ to its side. 

\myfigref{boxmap} shows how these operations are used to disambiguate {\small \tt Titanic} 
while exploiting the co-occurring mention {\small \tt Cameron} and the KG structure. 
We take the box for {\small \tt Cameron}, compute its neighborhood box, 
then intersect with the {\small \tt Titanic} box.
This intersection contains valid entities that can disambiguate {\small \tt Titanic} \emph{and} are connected to the entity for 
{\small \tt Cameron}. 
For the actual score of each such entity,
we take its distance to the center of the intersection box 
and convert it to a similarity score $Sim_{box}(\mathbf{m_i},\mathbf{e_{ij}})$. 
We then linearly combine this with the BERT-based similarity measure: 
$\beta_{box} Sim_{box}(\mathbf{m_i},\mathbf{e_{ij}}) + Sim_{cos}(\mathbf{m_i},\mathbf{e_{ij}})$, 
where $\beta_{box}$ is a hyper-parameter that adjusts the importance of the two scores. 
The approach described can be easily extended to more than two mentions.

\subsection{Model}
\label{sec:lnn-framework}
In this section, we describe how an EL algorithm composed of a disjunctive set of rules is reformulated into LNN representation for learning. 
\smallskip
\noindent \textbf{Entity Linking Rules} are a restricted form of FOL rules comprising of a set of Boolean predicates connected via logical operators: conjunction $(\wedge)$ and disjunction $(\vee)$. 
A Boolean predicate has the form $f_k > \theta$, where $f_k\in\mathcal{F}$ is one of the feature functions, and $\theta$ can be either a user provided or a learned threshold in $[0,1]$. 
\myfigref{fig:el-rules}(a) shows two example rules $R_1$ and $R_2$, where, for instance, $R_1(m_i,e_{ij})$ evaluates to {\tt \footnotesize True} if both the predicate $jacc(m_i,e_{ij})>\theta_1$ and $Ctx(m_i,e_{ij})>\theta_2$ are {\tt \footnotesize True}. 
Rules can be disjuncted together to form a larger EL algorithm, as the one shown in \myfigref{fig:el-rules}(b), 
which states that $Links(m_i,e_{ij})$ evaluates to {\tt \footnotesize True} if any one of its rules evaluates to {\tt \footnotesize True}.
The $Links$ predicate is meant to store high-quality links between mention and candidate entities that pass the conditions of at least one rule. The EL 
algorithm also acts as a scoring mechanism. In general, there are many ways in which scores can computed. In a baseline implementation (no learning), 
we use the scoring function in \myfigref{fig:el-rules}(c), where rw$_i$ denote manually assigned 
rule weights, while fw$_i$ are manually assigned feature weights. 

\begin{figure}[t]
    \centering
    \scalebox{.82}{\begin{tikzpicture}

\node at (0,0) (r1) {
\begin{minipage}{0.48\textwidth}
\small
$R_1(m_i,e_{ij})\leftarrow jacc(m_i,e_{ij}) > \theta_1\wedge Ctx(m_i,e_{ij}) > \theta_2 $
\end{minipage}
}; 

\node[above=2mm of r1] (l1) {\tt (a)\underline{EL Rules}};

\node[below=1mm of r1] (r2) {
\begin{minipage}{0.48\textwidth}
\small
$R_2(m_i,e_{ij})\leftarrow lev(m_i,e_{ij}) > \theta_3\wedge Prom(m_i,e_{ij}) > \theta_4 $
\end{minipage}
};

\node[below=1mm of r2] (l2) {\tt (b)\underline{EL Algorithm}};

\node[below=0mm of l2] (a1) {\begin{minipage}{0.4\textwidth}
\small
$Links(m_i,e_{ij})\leftarrow R_1(m_i,e_{ij}) \vee R_2(m_i,e_{ij})$

\end{minipage}
};
\node[below=1mm of a1] (l3) {\tt(c)\underline{\tt Scoring}};

\node[below=-0.2mm of l3] (s1) {\begin{minipage}{0.48\textwidth}
\small
$s(m_i,e_{ij})=$\\
\vspace{-1mm}
\begin{equation*}
    + \begin{pmatrix}
    rw_1\times ((fw_1\times jacc(m_i,e_{ij})\times(fw_2\times Ctx(m_i,e_{ij}))\\
    rw_2\times ((fw_3\times jacc(m_i,e_{ij})\times(fw_4\times Ctx(m_i,e_{ij}))
    \end{pmatrix}
   \end{equation*}
\end{minipage}
};

\end{tikzpicture}}
    \vspace{-1mm}
    \caption{Example of entity linking rules and scoring.}
    \vspace{-4mm}
    \label{fig:el-rules}
\end{figure}

An EL algorithm is an explicit and extensible description of the entity linking logic, 
which can be easily understood 
and manipulated by users. However, obtaining competitive performance to that of deep learning approaches such as BLINK~\cite{wu2019zero} requires a significant amount of manual effort to fine tune the thresholds $\theta_i$, 
the feature weights (fw$_i$) and the rule weights (rw$_i$). 

\noindent \textbf{LNN Reformulation.} 
To facilitate learning of the thresholds and weights in an EL algorithm, we map the Boolean-valued logic rules 
into the LNN formalism, where the LNN constructs -- LNN-$\vee$ (for logical OR) and LNN-$\wedge$ (for logical AND) -- 
allow for continuous real-valued numbers in $[0, 1]$. 
As described in ~\mysecref{sec:lnn-prelims}, LNN-$\wedge$ and LNN-$\vee$ are a weighted real-valued version of the classical logical operators, where a hyperparameter $\alpha$ is used as a proxy for $1$. Each LNN operator produces a value in $[0, 1]$ based on the values of the inputs, their weights and bias $\beta$. Both the weights and $\beta$ are 
learnable parameters. 
The score of each link is based on the score that the LNN operators give, with an added complication related to 
how we score the feature functions. To illustrate, for the EL rules in \myfigref{fig:el-rules}, the score of a link is computed as:

\smallskip
\begin{minipage}{0.47\textwidth}
\small
$s(m_i,e_{ij}) =$\\
\vspace{-1mm}
\begin{equation*}
    \text{LNN-}\vee \begin{pmatrix}
    
    \text{LNN-}\wedge \begin{pmatrix}
    TL(jacc(m_i,e_{ij}),\theta_1),\\
    TL(Ctx(m_i,e_{ij}),\theta_2)\\
    \end{pmatrix}
     ,\\ \\
      \text{LNN-}\wedge \begin{pmatrix}
    TL(lev(m_i,e_{ij}),\theta_3),\\
    TL(Prom(m_i,e_{ij}),\theta_4)\\
    \end{pmatrix}
      \end{pmatrix}
\end{equation*}
\end{minipage}

\smallskip
\noindent
Here the top-level LNN-$\vee$ represents the disjunction $R_1 \vee R_2$, while the two inner LNN-$\wedge$ capture 
the rules $R_1$ and $R_2$ respectively. 
For the feature functions with thresholds, a natural scoring mechanism would be to use 
${\tt \footnotesize score}(f > \theta) = f$ if $f > \theta$ else $0$, which filters out the candidates that do not satisfy the condition $f > \theta$, and gives a non-zero score for the candidates that pass the condition. 
However, since this is a step function which breaks the gradient flow through a neural network, 
we approximate it via a smooth function 
$TL(f,\theta)$ $= f\cdot \sigma(f-\theta)$, where $\sigma$ is Sigmoid function and $\theta$ is the learnable threshold that is generated using $\sigma$, i.e., $\theta = \sigma(\gamma)$, to ensure that it lies in $[0,1]$. 

\noindent \textbf{Training.} We train the LNN formulated EL rules over the labeled data and use a margin-ranking loss over all the candidates in $C_i$ to perform gradient descent. The loss function $\mathcal{L}(m_i,C_i)$ for mention $m_i$ and candidates set $C_i$ is defined as 

\vspace*{-4mm}
{\small 
\begin{align}
\sum_{e_{in}\in C_i\setminus\{e_{ip}\}}\hspace{-4mm}\max(0,-(s(m_i,e_{ip})-s(m_i,e_{in}))+\mu)\notag
\end{align}
}
Here, $e_{ip}\in C_i$ is a positive candidate, $C_i\setminus\{e_{ip}\}$ is the set of negative candidates, and $\mu$ is 
a margin hyper parameter. The positive and negative labels are obtained from the given labels $L_i$ (see ~\myfigref{fig:overview}).
\noindent \textbf{Inference.} Given mention $m_i$ and candidate set $C_i$, similar to training, we generate features for each mention-candidate pair ($m_i,e_{ij}$) in the feature generation step. We then pass them through the learned LNN network to 
obtain final scores for each candidate entity in $C_i$ as shown in \myfigref{fig:overview}. 
\section{Evaluation}
\vspace*{-1mm}
We first evaluate our 
approach w.r.t performance \& extensibility, interpretability and transferability. We also discuss the training and inference time. 
\begin{table}[t]
\vspace*{-1mm}
\begin{center}
\setlength{\tabcolsep}{3pt}
\footnotesize
\begin{tabular}{lcccc}
    \toprule
      \multirow{2}{*}{\bf Dataset}&\multicolumn{2}{c}{\bf Train} & \multicolumn{2}{c}{\bf Test}\\
      &|Q|&|E|&|Q|&|E|\\
      \midrule
     LC-QuAD 1.0{\scriptsize~\cite{10.1007/978-3-319-68204-4_22}}&4,000&6,823&1000&1,721\\
     QALD-9{\scriptsize~\cite{Usbeck20189thCO}}&408&568&150&174\\
     WebQSP\textsubscript{EL}{\scriptsize~\cite{li2020efficient}}&2974&3,237&1603&1,798\\
     \bottomrule
\end{tabular}
   \caption{Characteristics of the datasets.}
    \label{tab:datasets}
    \vspace*{-3mm}
\end{center}
\end{table}

\noindent\textbf{Datasets.} As shown in \mytabref{tab:datasets}, we consider 
three short-text QA datasets.
\noindent
LC-QuAD and QALD-9 are datasets comprising of questions (Q) over DBpedia together with their corresponding SPARQL queries. 
We extract entities (E) from SPARQL queries and manually annotate mention spans.
WebQSP\textsubscript{EL} dataset~\cite{li2020efficient} comprises of both mention spans and links 
to the correct entity. Since the target KG for WebQSP is Wikidata, we translate each Wikidata entity to its DBpedia counterpart using DBpedia Mappings\footnote{\url{http://mappings.dbpedia.org/}}. 
In addition, we discard mentions that link to DBpedia concepts (e.g., {\tt heaviest player} linked to {\tt dbo:Person}) and mentions $m_i$ with empty result (i.e., $C_i=\phi$) or all not-link labels (i.e, $\forall l_{ij}\in L_i, l_{ij}=0$)\footnote{Please check arXiv version for the datasets.}.

\begin{table*}[t]
\scriptsize
    \vspace*{-1mm}
    \centering
\begin{tabular}{lccccccccc}
    \toprule
    {\bf Model}&\multicolumn{3}{c}{\bf LC-QuAD}&\multicolumn{3}{c}{\bf QALD-9}&\multicolumn{3}{c}{\bf WebQSP\textsubscript{EL}}\\
    &Precision&Recall&F1&Precision&Recall&F1&Precision&Recall&F1\\
    \midrule
    BLINK&87.04&87.04&87.04 &89.14&89.14&89.14&92.15&92.05&92.10\\
    BERT&57.14&63.09&59.97&55.46&61.11&58.15&70.26&72.15&71.20\\
    BERTWiki&66.96&73.85&70.23&66.16&72.90&69.37&81.11&83.29&82.19\\
    Box&67.31&74.32&70.64&68.91&75.93&72.25&81.53&83.72&82.61\\
    \midrule
    {\em LogisticRegression}&87.04&86.83&86.93&84.73&84.73&84.73&83.39&83.33&83.36\\
     {\em LogisticRegression$_{BLINK}$}&90.50&90.30&90.40&88.94&88.94&88.94&89.33&89.28&89.31\\
     \midrule
    {\em RuleEL}&79.82&80.10&79.96&81.55&75.15&78.22&76.56&74.55&75.54\\
    {\em LogicEL}&86.68&86.48&86.58&83.05&83.05&83.05&82.60&82.58&82.59\\
    {\em LNN-EL} &87.74&87.54&87.64&88.52&88.52&88.52&85.11&85.05&85.08\\
    {\em LNN-EL$_{ens}$} &{\bf 91.10}&{\bf 90.90}&{\bf 91.00}&{\bf 91.38}&{\bf 91.38}&{\bf 91.38}&\textbf{92.17}&\textbf{92.08}&\textbf{92.12}\\
    \bottomrule
    \end{tabular}
    \vspace*{-2mm}
   \caption{Performance comparison of various baselines with our neuro-symbolic variants.}
    \label{tab:performance}
    \vspace*{-4mm}
\end{table*}

\smallskip
\noindent\textbf{Baselines.} We compare our approach to (1) {\em BLINK}~\cite{wu2019zero}, the current state-of-the-art on both short-text and long-text EL, 
(2) three BERT-based models - (a) {\em BERT}: both mention and candidate entity embeddings are obtained via BERT$_{base}$ pre-trained encoder, similar to~\cite{gillick2019learning}, 
(b) {\em BERTWiki}: mention embeddings are obtained from BERT$_{base}$, while candidate entity is from pretrained Wiki2Vec~\cite{yamada2020wikipedia2vec}, 
(c) {\em Box}: BERTWiki embeddings finetuned with Query2Box embeddings (see~\mysecref{sec:feature-functions}). In addition to the aforementioned black-box neural models, we also compare our approach to (3) two logistic regression models that use the same feature set as LNN-EL: {\em LogisticRegression} without BLINK and {\em LogisticRegression$_{BLINK}$} with BLINK.

Furthermore, we use the following variants of our approach: (4) {\em RuleEL}: a baseline rule-based EL approach with manually defined weights and thresholds, (5) {\em LogicEL}: a baseline approach built on RuleEL where only the thresholds are learnable, based on product $t$-norm (see~\mysecref{sec:lnn-prelims}), 
(6) {\em LNN-EL}: our core LNN-based method using non-embedding features plus SpaCy, 
and 
(7) {\em LNN-EL$_{ens}$}: an ensemble combining core LNN-EL with additional features 
from existing EL approaches, namely BLINK and Box (we consider Box, as it outperforms BERT and BERTWiki on all datasets). Detailed rule templates are provided in \myapdxref{apdx:lnnel-rules}.

\smallskip
\noindent\textbf{Setup.} All the baselines are trained for 30 epochs, except for BLINK which we use as a zero-shot approach. For BERT approaches, we use BERT\textsubscript{base} as pretrained model. We used two Nvidia V100 GPUs with 16GB memory each. We perform hyper-parameter search for margin $\mu$ and learning rates in the range $[0.6,0.95]$, $[10^{-5}, 10^{-1}]$ respectively.

\subsection{Results}
\noindent \textbf{Overall Performance.} As seen in \mytabref{tab:performance}, among logic-based approaches, LNN-EL outperforms LogicEL and RuleEL, showing 
that parameterized real-valued LNN learning is more effective than 
the non-parameterized version with $t$-norm (LogicEL) 
and the manually tuned RuleEL. Logistic regression models which also learn weights over features achieve competitive performance to LNN-EL models; however they lack the representation power that LNN-EL offer in the form of logical rules comprising of conjunctions and disjunctions. In other words, LNN-EL allows learning over a richer space of models that help in achieving better performance 
as observed in \mytabref{tab:performance}.

On the other hand, simple 
BERT-based approaches (BERT, BERTWiki, Box) that are trained on the QA datasets 
underperform the logic-based approaches, which incorporate finer-grained features. 
BLINK (also a BERT-based approach, but trained on the entire Wikipedia) is 
used as zero-shot approach and achieves SotA performance (when not counting the LNN-EL variants). 
The core LNN-EL version is competitive with BLINK on LC-QuAD and QALD-9, despite being a rule-based approach. Furthermore, LNN-EL$_{ens}$, which combines the core LNN-EL with both BLINK and Box features, easily beats BLINK on LC-QuAD and QALD-9 and slightly on WebQSP\textsubscript{EL}.

\mytabref{tab:topk-performance} shows the Recall@$k$ performance of LNN-EL against the BLINK model. Both LNN-EL and LNN-EL$_{ens}$ have better Recall@$k$ performance against BLINK on LC-QuAD and QALD-9 datasets, however BLINK's Recall@$k$ achieves a slightly better performance for WebQSP\textsubscript{EL} dataset.

\begin{table}[tbh]
\scriptsize
\setlength{\tabcolsep}{5pt}
    \centering
    \begin{tabular}{llccc}
         \toprule
        Dataset&Model&R@5&R@10&R@64\\
        \midrule
        \multirow{2}{*}{LC-QuAD}
        &BLINK&94.69&96.01&96.92\\
        &LNN-EL &93.66&94.39&97.56\\
        &LNN-EL$_{ens}$ &97.07&97.20&97.68\\
        \midrule
        \multirow{2}{*}{QALD-9}
        &BLINK&93.39&93.39&94.29\\
        &LNN-EL &92.72&95.94&98.04\\
        &LNN-EL$_{ens}$&94.63&94.63&95.48\\
        \midrule
        \multirow{2}{*}{WebQSP\textsubscript{EL}}
        &BLINK&97.40&97.64&98.61\\
        &LNN-EL  &93.54&95.12&96.59\\
        &LNN-EL$_{ens}$&96.34&96.59&96.95\\
        \bottomrule
    \end{tabular}
    \caption{Recall@$k$ performance of LNN-EL models}
    \vspace*{-1mm}
    \label{tab:topk-performance}
\end{table}

\noindent \textbf{Extensibility.}
Here, we inspect empirically how a multitude of EL features coming from various black-box approaches 
can be combined in a principled way with LNN-EL, often leading to an overall 
better performance than the individual approaches. A detailed ablation study of the core LNN-EL version can be found in \myapdxref{apdx:ablation}.
As seen in \mytabref{tab:extensibility}, approaches like BERTWiki and Box which in isolation underperform compared to LNN-EL, help boost the latter's performance if they are included as predicates.  Similarly, LNN-EL which has comparable performance to BLINK, can accommodate the latter's score to produce better performance (see LNN-EL$_{+{\tt BLINK}}$). We also note that adding features is not a guarantee  to improve performance, as LNN-EL$_{ens}$ (which includes both BLINK and Box) slightly underperforms LNN-EL$_{+{\tt BLINK}}$ on WebQSP\textsubscript{EL}. For such cases, the interpretability of LNN-EL (discussed next) can help users select the right features based on their relative importance.

\begin{table}[t]
\scriptsize
\setlength{\tabcolsep}{1.5pt}
    \centering
    \begin{tabular}{lccccc}
         \toprule
         Dataset& LNN-EL& LNN-EL& LNN-EL& LNN-EL&LNN-EL$_{ens}$\\
         & & ~~+BLINK&~~+BERTWiki&~~+Box&\\
         \midrule
         LC-QuAD&87.64&90.24&88.23&89.05&{\bf 91.00}\\
         QALD-9&88.52&90.96&86.41&88.52&\textbf{91.38}\\
         WebQSP\textsubscript{EL}&85.08&{\bf 92.32}&91.70&91.44&92.12\\
         \bottomrule
    \end{tabular}
    \caption{F1 scores of LNN-EL with additional features coming from various black-box EL approaches.} 
    \vspace*{-4mm}
    \label{tab:extensibility}
    
\end{table}

\begin{figure}
    \centering
    \scriptsize
    \begin{tikzpicture}
\node at (0,0) (lnnens) {\begin{minipage}{0.25\textwidth}
\begin{tikzpicture}
\node[draw, rounded corners] at (0,0) (mainor) {$\vee$};
\node[below = 5mm of mainor] (d1) {};
\node[draw, rounded corners, left = 5mm of d1] (blink) {BLINK};
\node[draw, rounded corners, right = 5mm of d1] (dbpor) {$\vee$};
\node[draw, rounded corners, below = 5mm of dbpor] (sim) {Sim$_\wedge$};
\node[draw, rounded corners, left = 3mm of sim] (ctx) {Ctx};
\node[draw, rounded corners, right = 3mm of sim] (type) {Type};
\node[below = 5mm of sim] (d2) {};
\node[draw, rounded corners,left = 2mm of d2] (simor) {$\vee$};
\node[draw, rounded corners,right = 2mm of d2] (prom) {$Prom_\theta$};
\node[below = 5mm of simor] (d3) {\ldots};
\node[draw, rounded corners,left = 2mm of d3] (jacc) {$Lev_\theta$};
\node[draw, rounded corners,right = 2mm of d3] (box) {$Box_\theta$};

\path (mainor) edge node[rotate=0,above=-0.8mm,fill=white,pos=0.7] {$0.89$} (blink);
\path (mainor) edge node[rotate=0,above=-0.6mm,fill=white,pos=0.7] {$0.26$} (dbpor);

\path (dbpor) edge node[rotate=0,above=-0.8mm,fill=white,pos=0.7,inner sep=0.3mm] {$0.22$} (ctx);
\path (dbpor) edge node[rotate=0,above=-1.5mm,fill=white,pos=0.5,inner sep=0.3mm] {$0.72$} (sim);
\path (dbpor) edge node[rotate=0,above=-0.8mm,fill=white,pos=0.7,inner sep=0.3mm] {$0.19$} (type);
\path (sim) edge
node[rotate=0,above=-0.8mm,fill=white,pos=0.5,inner sep=0.3mm] {$0.18$} (simor);
\path (sim) edge node[rotate=0,above=-0.8mm,fill=white,pos=0.5,inner sep=0.3mm] {$0.81$} (prom);
\path (simor) edge node[rotate=0,above=-0.8mm,fill=white,pos=0.5,inner sep=0.3mm] {$0.16$} (jacc);
\path (simor) edge node[rotate=0,above=-0.8mm,fill=white,pos=0.5,inner sep=0.3mm] {$0.69$} (box);
\end{tikzpicture}

\end{minipage}
};
\node at (4,0) (lnnelblink) {\begin{minipage}{0.25\textwidth}
\begin{tikzpicture}
\node[draw, rounded corners] at (0,0) (mainor) {$\vee$};
\node[below = 5mm of mainor] (d1) {};
\node[draw, rounded corners, left = 5mm of d1] (blink) {BLINK};
\node[draw, rounded corners, right = 5mm of d1] (dbpor) {$\vee$};
\node[draw, rounded corners, below = 5mm of dbpor] (sim) {Sim$_\wedge$};
\node[draw, rounded corners, left = 3mm of sim] (ctx) {Ctx};
\node[draw, rounded corners, right = 3mm of sim] (type) {Type};
\node[below = 5mm of sim] (d2) {};
\node[draw, rounded corners,left = 2mm of d2] (simor) {$\vee$};
\node[draw, rounded corners,right = 2mm of d2] (prom) {$Prom_\theta$};
\node[below = 5mm of simor] (d3) {\ldots};
\node[draw, rounded corners,left = 2mm of d3] (jacc) {$Lev_\theta$};

\path (mainor) edge node[rotate=0,above=-0.8mm,fill=white,pos=0.7] {$1.01$} (blink);
\path (mainor) edge node[rotate=0,above=-0.6mm,fill=white,pos=0.7] {$0.42$} (dbpor);

\path (dbpor) edge node[rotate=0,above=-0.8mm,fill=white,pos=0.7] {$0.55$} (ctx);
\path (dbpor) edge node[rotate=0,above=-1.5mm,fill=white,pos=0.5,inner sep=0.3mm] {$0.35$} (sim);
\path (dbpor) edge node[rotate=0,above=-0.8mm,fill=white,pos=0.6,inner sep=0.3mm] {$0.47$} (type);
\path (sim) edge node[rotate=0,above=-0.8mm,fill=white,pos=0.5,inner sep=0.3mm] {$0.94$} (simor);
\path (sim) edge node[rotate=0,above=-0.8mm,fill=white,pos=0.5,inner sep=0.3mm] {$0.84$} (prom);
\path (simor) edge node[rotate=0,above=-0.8mm,fill=white,pos=0.5,inner sep=0.3mm] {$0.10$} (jacc);
\end{tikzpicture}

\end{minipage}
};

\end{tikzpicture}
    \caption{Feature weights of two models LNN-EL$_{ens}$(left) and LNN-EL$_{+{\tt BLINK}}$ (right) on WebQSP\textsubscript{EL}}
    \label{fig:lnn-viz}
    \vspace*{-2mm}
\end{figure}
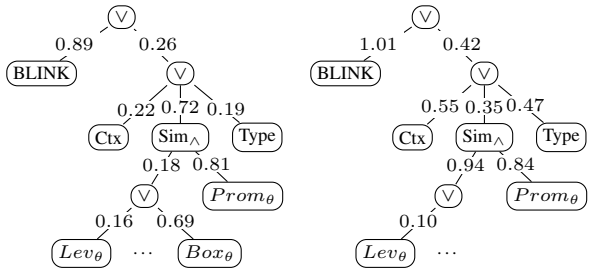

\smallskip
\noindent \textbf{Interpretability.} Unlike black-box models, rule-based approaches provide the capability to inspect the model, specifically on how the features impact performance. This inspection can help in dropping or adjusting features that are detrimental. 
For instance, consider 
our case of 
LNN-EL$_{+{\tt BLINK}}$ and LNN-EL$_{ens}$ trained on WebQSP\textsubscript{EL} dataset, 
where we observed 
that LNN-EL$_{ens}$'s performance is inferior to LNN-EL$_{+{\tt BLINK}}$ even though the former model has more features. 
A human expert can find insights into this behavior 
by looking at the 
feature weights in each model. 
In \myfigref{fig:lnn-viz} (left), the disjunction tree with the $Box$ feature is given a low 
weight of $0.26$, thus discounting some of the other useful features in the same tree. 
Removal of the Box feature leads to a re-weighting of the features in the model; 
the modified disjunction tree (\myfigref{fig:lnn-viz} (left)) has now a weight of $0.42$.
Such visualization can help the rule designer to judiciously select features to combine towards building a performant model. 

\noindent \textbf{Transferability.} To study the transferability aspect, 
we train LNN-EL on one dataset and evaluate the model on the other two, without any finetuning. 
We use the core LNN-EL variant for this, but similar properties hold for the other variants. 
\mytabref{tab:transferability} shows F1 scores on different train-test configurations, with diagonal (underlined numbers) denoting the F1 score when trained and tested on the same dataset. 
We observe that LNN-EL transfers reasonably well, even in cases where training is done on a very small dataset. For example, when we transfer from QALD-9 (with only a few hundred questions to train) to WebQSP\textsubscript{EL}, we obtain an F1-score of 
$83.06$ which is within 2 percentage points of the F1-score when trained directly on WebQSP\textsubscript{EL}. We remark that the zero-shot BLINK by design has very good transferability 
and achieves F1 scores of 87.04, 89.14, 92.10 on LC-QuAD, QALD-9, WebQSP\textsubscript{EL} respectively. 
However, BLINK is trained on the entire Wikipedia, while LNN-EL needs much less data 
to achieve reasonable transfer performance.    

\begin{table}[t]
\scriptsize
    \centering
    \begin{tabular}{l|ccc}
       \toprule
         \multirow{2}{*}{Train}&\multicolumn{3}{c}{Test}  \\
         & LC-QuAD&QALD-9&WebQSP\textsubscript{EL}\\
         \midrule
         LC-QuAD&\underline{87.64}&86.41&78.90\\
         QALD-9&85.58&\underline{88.52}&83.06\\
         WebQSP\textsubscript{EL}&80.95&87.25&\underline{85.08}\\
       \bottomrule
       
    \end{tabular}
    \caption{F1 scores of LNN-EL in transfer settings.}
    \label{tab:transferability}
    \vspace{-3mm}
\end{table}

\eat{
\begin{table}[t]
\scriptsize
 \centering
 \scalebox{1.0}{
    \begin{tabular}{llccc}
         \toprule
         Test&Train&Precision&Recall&F1\\
         \midrule
         \multirow{3}{*}{LC-QuAD}
         &{LC-QuAD}&87.74&87.54&87.64\\
         &{QALD-9}&85.68&85.48&85.58\\
         &{WebQSP}&81.06&80.85&80.95\\
         \midrule

         \multirow{3}{*}{WebQSP}
         &{WebQSP}&85.11&85.05&85.08\\
         &{QALD-9}&83.09&83.04&83.06\\
         &{LC-QuAD}&78.93&78.87&78.90\\
         \midrule

         \multirow{3}{*}{QALD-9}
         &{QALD-9}&88.52&88.52&88.52\\
         &{WebQSP}&87.25&87.25&87.25\\
         &{LC-QuAD}&86.41&86.41&86.41\\
         \bottomrule
    \end{tabular}
    }
    \caption{F1 performance of LNN-EL }
    \label{tab:transferability}
\end{table}
}

\medskip
\medskip
\noindent \textbf{Runtime Analysis.} We study the efficiency of LNN-EL$_{ens}$ across three aspects: 1) candidate \& feature generation, 2) training, and 3) inference. Candidate \& feature generation involve using the DBpedia lookup API
to obtain candidates for each mention, pruning non-entity candidates (i.e., categories, disambiguation links, etc.), obtaining any missing descriptions for candidates using SPARQL endpoint, and finally generating feature vectors for each mention-candidate pair using the feature functions described in \mysecref{sec:feature-functions}. The generated features for the train and test data are then used, respectively, to train and test the LNN-EL models.  The number of parameters in an LNN-EL model is linearly proportional to the combined number of disjunctions and conjunctions, which typically is in the order of few 10s. For example, LNN-EL$_{ens}$ comprises of 72 parameters, which is several orders of magnitude smaller than in neural black box models such as BLINK. \mytabref{tab:runtimeanalysis} provides the time (in seconds) taken per question for candidate \& feature generation, as well as 5-run average training and inference time per epoch. 

\begin{table}[]
    \centering
    \scriptsize
    \begin{tabular}{llll}
         \toprule
         &Candidate \& feature  & Training & Inference  \\
         &generation&per epoch& per epoch\\
         \midrule
         QALD-9&26.21&0.010&0.009\\
         LC-QuAD&33.05&0.010&0.013\\
         WebQSP\textsubscript{EL}&19.80&0.009&0.012\\
         \bottomrule
    \end{tabular}
    \caption{Time per question for candidate \& feature generation, along with train and inference time per question for LNN-EL$_{ens}$. All numbers are in seconds.}
    \vspace{-7mm}
    \label{tab:runtimeanalysis}
    
\end{table}

\vspace{-3mm}
\section{Conclusions}
\vspace{-1.5mm}
We introduced LNN-EL, a neuro-symbolic approach for entity linking on short text. Our approach complements 
human-given rule templates through neural learning and achieves competitive performance against SotA black-box neural models, while exhibiting interpretability and transferability without requiring a large amount of labeled data. While LNN-EL provides an extensible framework where one can easily add and test new features in existing rule templates,  currently this is done manually. A future direction is to automatically learn the rules with the optimal combinations of features.

\vspace{-1.5mm}
\section*{Acknowledgements}
\vspace{-1.5mm}
We thank Ibrahim Abdelaziz, Pavan Kapanipathi, Srinivas Ravishankar, Berthold Reinwald, Salim Roukos and anonymous reviewers for their valuable inputs and feedback.

\bibliography{acl2021}
\bibliographystyle{acl_natbib}

\clearpage
\appendix
\section{Appendix}
\label{sec:supplemental}

\subsection{$t$-norm and $t$-conorm}
\label{apdx:lnn}
While linear classifiers, decision lists/trees may also be considered interpretable, rules expressed in first-order logic (FOL) form a much more powerful, closed language that offer semantics clear enough for human interpretation and a larger range of operators facilitating the expression of richer models. To learn these rules, neuro-symbolic AI substitutes conjunctions (disjunctions) with differentiable $t$-norms ($t$-conorms) \citep{esteva:fuzzysetssyst01}. However, since it does not have any learnable parameters, 
this behavior cannot be adjusted, which limits how well it can model the data. For example, while linear classifiers such as logistic regression can only express a (weighted) sum of features which is similar to logic's disjunction ($\vee$) operator, logic also contains other operators including, but not limited to, conjunction ($\wedge$), and negation ($\neg$). 

As opposed to inductive logic programming \citep{muggleton:ilpw96} and statistical relational learning \citep{getoor:book}, neuro-symbolic AI utilizes neural networks to learn rules. Towards achieving this, the first challenge to overcome is that classical Boolean logic is non-differentiable and thus, not amenable to gradient-based optimization (e.g., backpropagation). To address this, neuro-symbolic AI substitutes conjunctions (disjunctions) with differentiable $t$-norms ($t$-conorms) \citep{esteva:fuzzysetssyst01}. For example, product $t$-norm, used in multiple neuro-symbolic rule-learners \citep{evans:jair18,yang:nips17}, is given by
$x \wedge y \equiv xy$,
where $x,y \in [0,1]$ denote input features in real-valued logic. Product $t$-norm agrees with Boolean conjunction at the extremities, i.e., when $x,y$ are set to $0$ ({\tt false}) or $1$ ({\tt true}). However, when $x,y \in [0,1] \setminus \{0,1\}$, its behavior is governed by the product function. More importantly, since it does not have any learnable parameters, 
this 
behavior cannot be adjusted, 
which limits how well it can model the data.

\subsection{Ablation Study}
\label{apdx:ablation}
To understand the roles of eac rule in LNN-EL, we also conduct ablation study on the largest benchmark dataset LC-QuAD (see Table \ref{sec:ablation}). We observe that Context is the most performant rule alone. Although PureName rule is behind the other two alone, PureName + Context improves the performance of Context by 1\%. Meanwhile, Context + Type only improves Context's performance by 0.05\%. Interestingly, the combination of three rules performs slightly worse than PureName + Context by 0.35\%. These results show that Type rule is less important among the three rules. To be consistent with the RuleEL system, we apply ``PureName + Context + Type'' setting for LNN-EL in our experiments. 

\begin{table}[tbh]
\footnotesize
    \centering
    \begin{tabular}{lccc}
         \toprule
        Dataset&Precision & Recall & F1\\
        \midrule
        PureName &76.03&75.83&75.93\\
        ~~~+ Context &88.09&87.89&87.99\\
        ~~~~~~+ Type &87.74&87.54&87.64\\
        ~~~+ Type &81.46&81.26&81.36\\
        \midrule
        Context &87.04&86.83&86.93\\
        ~~~+ Type &87.09&86.88&86.98\\
        \midrule
        Type &87.04&86.83&86.93\\
        \bottomrule
    \end{tabular}
    \caption{LNN-EL Ablation Analysis on LC-QuAD}
    \label{sec:ablation}
\end{table}

Additionally, we also show the transferability of LR in Table \ref{tab:lr-transferability}. This must be compared with the corresponding LNN-EL results in the earlier Table \ref{tab:transferability}. In particular, we observe that LNN-EL outperforms LR in 4 out of 6 transferability tests, demonstrating that LNN-EL has superior transferability. 
\begin{table}[t]
\scriptsize
    \centering
    \begin{tabular}{l|ccc}
       \toprule
         \multirow{2}{*}{Train}&\multicolumn{3}{c}{Test}  \\
         & LC-QuAD&QALD-9&WebQSP\textsubscript{EL}\\
         \midrule
         LC-QuAD&\underline{86.93}&84.73&76.72\\
         QALD-9&87.14&\underline{84.73}&80.03\\
         WebQSP\textsubscript{EL}&83.42&86.83&\underline{83.59}\\
       \bottomrule
       
    \end{tabular}
    \caption{F1 scores of LR in transfer settings.}
    \vspace{-3mm}
    \label{tab:lr-transferability}
\end{table}

\subsection{LNN-EL Rules}
\label{apdx:lnnel-rules}
In our experiments, we explore the following modules, implemented in PyTorch. 
\smallskip

\noindent \textbf{Name Rule:}

\resizebox{.9\linewidth}{!}{
  \begin{minipage}{\linewidth}
\begin{align*}
      R_{name} \leftarrow &~~[f_{jacc}(m_i,e_{ij}) > \theta_1 \vee f_{lev}(m_i,e_{ij}) > \theta_2\notag\\
      & \vee f_{jw}(m_i,e_{ij}) > \theta_3  \vee f_{spacy}(m_i,e_{ij}) > \theta_4]\notag\\
      &\wedge f_{prom}(m_i,e_{ij})\notag
\end{align*}
  \end{minipage}
}

\vskip 0.2in
\noindent \textbf{Context Rule:}

\resizebox{.9\linewidth}{!}{
  \begin{minipage}{\linewidth}
\begin{align*}
      R_{ctx} \leftarrow &~~[f_{jacc}(m_i,e_{ij}) > \theta_1 \vee f_{lev}(m_i,e_{ij}) > \theta_2\notag\\
      & \vee f_{jw}(m_i,e_{ij}) > \theta_3  \vee f_{spacy}(m_i,e_{ij}) > \theta_4]\notag\\
      &\wedge f_{ctx}(m_i,e_{ij}) > \theta_5\notag\\
      &\wedge f_{prom}(m_i,e_{ij})\notag
\end{align*}
  \end{minipage}
}

\vskip 0.2in
\noindent \textbf{Type Rule:}

\resizebox{.9\linewidth}{!}{
  \begin{minipage}{\linewidth}
\begin{align*}
      R_{type} \leftarrow &~~[f_{jacc}(m_i,e_{ij}) > \theta_1 \vee f_{lev}(m_i,e_{ij}) > \theta_2\notag\\
      & \vee f_{jw}(m_i,e_{ij}) > \theta_3  \vee f_{spacy}(m_i,e_{ij}) > \theta_4]\notag\\
      &\wedge f_{type}(m_i,e_{ij}) > \theta_5\notag\\
      &\wedge f_{prom}(m_i,e_{ij})\notag
\end{align*}
  \end{minipage}
}

\vskip 0.2in
\noindent \textbf{Blink Rule:}

\resizebox{.9\linewidth}{!}{
  \begin{minipage}{\linewidth}
\begin{align*}
      R_{blink} \leftarrow &~~[f_{jacc}(m_i,e_{ij}) > \theta_1 \vee f_{lev}(m_i,e_{ij}) > \theta_2\notag\\
      & \vee f_{jw}(m_i,e_{ij}) > \theta_3  \vee f_{spacy}(m_i,e_{ij}) > \theta_4]\notag\\
      &\wedge f_{blink}(m_i,e_{ij})
\end{align*}
  \end{minipage}
}

\vskip 0.2in
\noindent \textbf{Box Rule:}

\resizebox{.9\linewidth}{!}{
  \begin{minipage}{\linewidth}
\begin{align*}
      R_{box} \leftarrow &~~[f_{jacc}(m_i,e_{ij}) > \theta_1 \vee f_{lev}(m_i,e_{ij}) > \theta_2\notag\\
      & \vee f_{jw}(m_i,e_{ij}) > \theta_3  \vee f_{spacy}(m_i,e_{ij}) > \theta_4]\notag\\
      & \vee f_{box}(m_i,e_{ij}) > \theta_5
\end{align*}
  \end{minipage}
}

\vskip 0.2in
\noindent \textbf{BERT Rule:}

\resizebox{.9\linewidth}{!}{
  \begin{minipage}{\linewidth}
\begin{align*}
      R_{bert} \leftarrow &~~[f_{jacc}(m_i,e_{ij}) > \theta_1 \vee f_{lev}(m_i,e_{ij}) > \theta_2\notag\\
      & \vee f_{jw}(m_i,e_{ij}) > \theta_3  \vee f_{spacy}(m_i,e_{ij}) > \theta_4]\notag\\
      & \vee f_{bert}(m_i,e_{ij}) > \theta_5
\end{align*}
  \end{minipage}
}

\smallskip
\noindent \textbf{LNN-EL:}

\resizebox{.9\linewidth}{!}{
  \begin{minipage}{\linewidth}
\begin{align*}
      R_{LNN-EL} \leftarrow & R_{name} \vee R_{ctx} \vee R_{type} 
\end{align*}
  \end{minipage}
}

\medskip
\noindent \textbf{LNN-EL$_{+{\tt BLINK}}$:}

\resizebox{.9\linewidth}{!}{
  \begin{minipage}{\linewidth}
\begin{align*}
      R_{LNN-EL_{+{\tt BLINK}}} \leftarrow & R_{LNN-EL} \vee R_{blink}
\end{align*}
  \end{minipage}
}

\medskip
\noindent \textbf{LNN-EL$_{ens}$:}

\resizebox{.9\linewidth}{!}{
  \begin{minipage}{\linewidth}
\begin{align*}
      R_{LNN-EL_{ens}} \leftarrow & R_{LNN-EL} \vee R_{blink} \vee R_{box}
\end{align*}
  \end{minipage}
}

\end{document}